\documentclass{article}     

\usepackage{arxiv}
\usepackage[utf8]{inputenc} 
\usepackage[T1]{fontenc}    
\usepackage{hyperref}       
\usepackage{url}            
\usepackage{booktabs}       
\usepackage{amsfonts}       
\usepackage{nicefrac}       
\usepackage{microtype}      
\usepackage{lipsum}
\usepackage{graphicx}
\usepackage{caption}
\usepackage{subcaption}
\usepackage{multirow}
\usepackage{amsmath,amsfonts}
\usepackage{tensor}
\usepackage{multirow,multicol}
\usepackage{orcidlink}
\usepackage[capitalise]{cleveref}
\usepackage{wrapfig}
\title{MagiClaw: A Dual-Use, Vision-Based Soft Gripper for Bridging the Human Demonstration to Robotic Deployment Gap}
\author{
    Tianyu Wu, Xudong Han, Haoran Sun, Zishang Zhang, Bangchao Huang \\
    Design + Learning Research Group\\
    Southern University of Science and Technology\\
    \AND
    Chaoyang Song \\
    asRobotics\\
    \texttt{songcy@ieee.org} \\
	\And
	Fang Wan\thanks{Corresponding Author.} \\
	SUSTech\\
	\texttt{wanfang@ieee.org} \\
}
\begin{document}
\maketitle
\begin{abstract}

    The transfer of manipulation skills from human demonstration to robotic execution is often hindered by a ``domain gap'' in sensing and morphology. This paper introduces MagiClaw, a versatile two-finger end-effector designed to bridge this gap. MagiClaw functions interchangeably as both a handheld tool for intuitive data collection and a robotic end-effector for policy deployment, ensuring hardware consistency and reliability. Each finger incorporates a Soft Polyhedral Network (SPN) with an embedded camera, enabling vision-based estimation of 6-DoF forces and contact deformation. This proprioceptive data is fused with exteroceptive environmental sensing from an integrated iPhone, which provides 6D pose, RGB video, and LiDAR-based depth maps. Through a custom iOS application, MagiClaw streams synchronized, multi-modal data for real-time teleoperation, offline policy learning, and immersive control via mixed-reality interfaces. We demonstrate how this unified system architecture lowers the barrier to collecting high-fidelity, contact-rich datasets and accelerates the development of generalizable manipulation policies. Please refer to the iOS app at \url{https://apps.apple.com/cn/app/magiclaw/id6661033548} for further details. 
    
\end{abstract}
\section{Introduction}
\label{sec:Intro}

    The success of modern robot learning paradigms, from Learning from Demonstration (LfD) \cite{argall2009survey, billard2008robot} to offline reinforcement learning, is fundamentally dependent on the quality and richness of the underlying data \cite{barekatain2024a}. For contact-rich manipulation tasks, robust policies require more than just kinematic trajectories; they demand a holistic understanding of interaction forces, tactile feedback, and environmental context \cite{jin2024visual, liu2024forcemimic}. Consider a human deftly handling a delicate object: the action is a symphony of precise motion, modulated forces, and continuous tactile adjustments \cite{billard2019trends}. Replicating such skills requires capturing this multi-modal information stream in its entirety.
    
    However, existing data collection methodologies present significant challenges. First, they often rely on a patchwork of disparate, expensive sensors—such as external motion capture systems, wrist-mounted force/torque sensors, and complex tactile skins \cite{lee2000tactile, meribout2024tactile}—resulting in cumbersome and costly setups. This high barrier to entry limits the scale and diversity of data collection efforts \cite{levine2018learning}. Second, and more critically, a persistent \textbf{domain gap} exists between the human demonstrator and the robotic learner \cite{ravichandar2020recent}. Data is often collected using one set of hardware (e.g., an instrumented glove) and deployed on a robot with entirely different sensor suites and end-effector morphology. This mismatch necessitates complex domain adaptation techniques and is a primary reason why policies trained on demonstration data often fail to generalize to physical hardware \cite{zare2024a}.
    
    \begin{figure*}[b]
        \centering
        \includegraphics[width=1.0\linewidth]{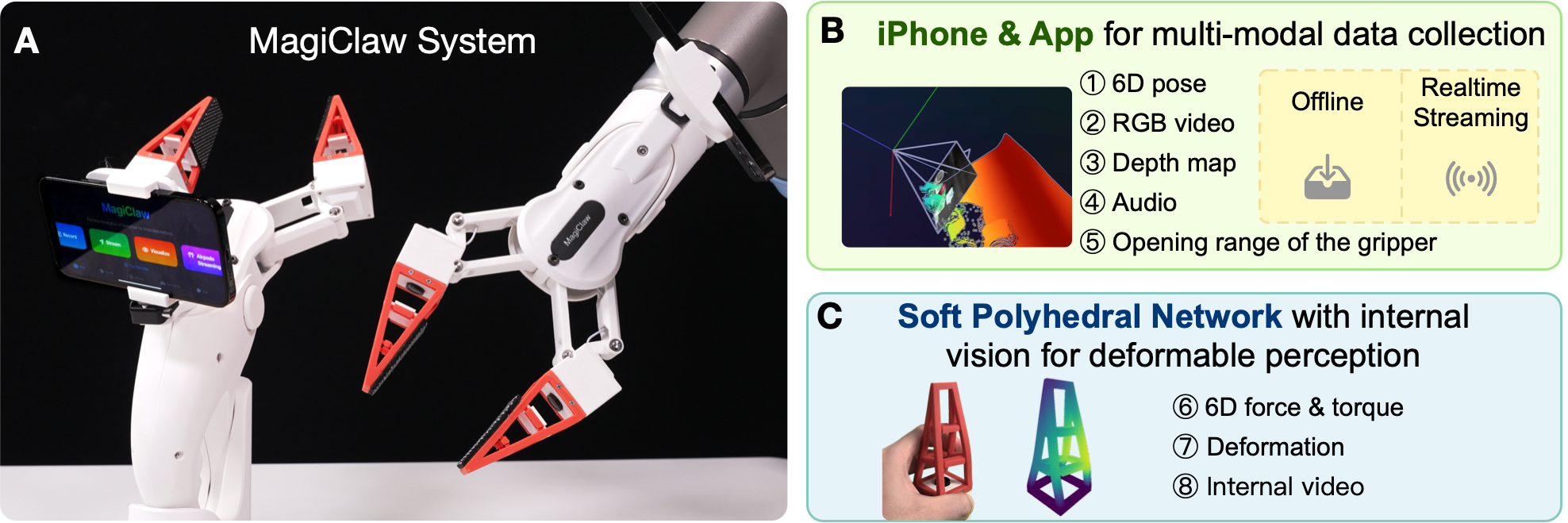}
        \caption{\textbf{Overview of the MagiClaw System.}
        (A) MagiClaw's dual-purpose design: a hand-held device for intuitive demonstration, data collection, and an identical end-effector mounted on a robotic arm for policy deployment.
        (B) An integrated iPhone provides exteroceptive sensing (6D pose, RGB, Depth) and a user interface for data management.
        (C) The Soft Polyhedral Network (SPN) fingertip contains an internal camera for vision-based proprioceptive sensing of 6D forces, torque, and local deformation.}
        \label{fig:MagiClaw}
    \end{figure*}
    
    To address these challenges, we present \textbf{MagiClaw}, a unified hardware platform designed to seamlessly bridge the gap from human demonstration to robotic deployment. MagiClaw is a dual-purpose, two-fingered gripper that merges three key innovations:
    \begin{enumerate}
        \item \textbf{Unified Hardware Form Factor:} The exact same MagiClaw device can be used as a hand-held tool for human demonstration or mounted on a robot arm for autonomous execution. This hardware consistency minimizes the sensor and morphological domain gap, facilitating direct policy transfer.
        \item \textbf{Vision-Based Proprioceptive Fingertips:} Each finger integrates a \emph{Soft Polyhedral Network} (SPN) \cite{liu2024proprioceptive} with an embedded miniature camera. This novel design enables \emph{visuotactile perception}, inferring 6-DoF forces, torque, and high-resolution contact deformation from the distortion of the internal lattice structure, thereby obviating the need for costly external force sensors.
        \item \textbf{Integrated Multi-Modal Exteroception:} An attached iPhone leverages its powerful sensor suite (LiDAR, RGB cameras, IMU) and ARKit framework \cite{apple2025arkit} to provide synchronized, rich environmental context, including gripper pose, depth maps, and high-resolution video.
    \end{enumerate}
    
    Our primary contribution is an integrated system that fundamentally streamlines the collection of holistic, contact-centric data for robot learning. By fusing proprioceptive force/tactile data from the fingertips with exteroceptive visual and spatial data from a commodity smartphone, MagiClaw offers a low-cost, powerful, and user-friendly solution for both teleoperation and autonomous policy development. We posit that by democratizing access to such high-fidelity, multi-modal data, MagiClaw can serve as a catalyst for developing more robust and generalizable manipulation skills, advancing the pursuit of universal action embodiment in robotics.
    
\section{The MagiClaw Gripper System}
\label{sec:MagiClaw}

    MagiClaw is designed to bridge the gap between human demonstration and autonomous robotic manipulation through a \emph{unified}, \emph{dual-purpose} gripper system. As shown in Fig.~\ref{fig:MagiClaw}, it can function either as a \textbf{hand-held tool} for collecting multi-modal data in human-guided demonstrations or as a \textbf{robotic end-effector} mounted onto an industrial or collaborative robot arm. By sharing identical hardware and sensor layouts in these two modes, MagiClaw minimizes sensor disparities that often hinder the transfer of learned skills from humans to robots.

\subsection{Engineering Highlights}

    \paragraph{Mechanical Design as a Dual-Purpose EOAgenT} 
    Fig.~\ref{fig:MagiClaw}A shows the overall design, which is inspired by widely adopted industrial solutions such as OnRobot's RG2 gripper. However, we completely redesigned the entire mechanical system for use in robot learning scenarios, ensuring it is suitable for dual-purpose usage by both human operators and robotic arms.

    The \textit{\textbf{base design}} features a parallel four-bar gripper mechanism, actuated by a back-driable motor, with a detachable iPhone mount, and two omni-adaptive fingertips with an in-finger miniature camera capable of Vision-based Deformable Perception.

    \textit{\textbf{Two variations}} are currently available, including a \textit{Hand-Held Mode} for data collection and an \textit{End-Effector mode} for robotic manipulation, formulating an End-of-Arm-Agent (EOAgenT) system.
    \begin{itemize}
        \item \textbf{Hand-Held Mode}
            \begin{itemize}
                \item \textbf{Ergonomic Handle and Trigger}: When used in \emph{hand-held} mode (Fig.~\ref{fig:MagiClaw}A), a molded handle accommodates the user's grip, and a trigger mechanism directly manipulates the finger openings. This setup enables an operator to perform everyday tasks, such as lifting, placing, sliding, or rotating objects, just as they would with a normal tool. Meanwhile, MagiClaw's onboard sensors continuously log force, pose, and environmental context without impeding the user's natural motions.
                \item \textbf{Live Data Capture for Demonstrations}: Because the same system can later be mounted on a robot, the hand-held demonstration data (trajectories, forces, tactile events) directly translate into robotic replay or training sets for \emph{imitation learning}. Operators can also leverage real-time visual or force feedback to refine their demonstrations in real-time. This approach significantly \emph{lowers the entry barrier} to collecting rich multi-modal data, even outside specialized lab environments.
            \end{itemize}
        \item \textbf{End-Effector Mode}
            \begin{itemize}
                \item \textbf{Mounting and Interface}: In robotic deployments (Fig.~\ref{fig:MagiClaw}B), the handle and trigger assembly can be detached, and the same mechanical finger unit is secured onto a standard robotic flange (e.g., an ISO 9409-1 mount). The iPhone remains attached to the gripper, maintaining the same sensing configurations. A single cable or wireless link connects the microcontroller to the robot's main controller, issuing commands that open/close or apply force.
                \item \textbf{Closing the Demonstration-to-Deployment Loop}: This \emph{dual-use design} is key to minimizing discrepancies between human-collected data and final robotic execution. By ensuring sensor placement, geometry, and compliance remain the same, MagiClaw helps learned policies replicate the human-demonstrated strategies more accurately. Tasks initially \emph{taught} to the robot in hand-held mode---like gently grasping a delicate object or manipulating flexible packaging---can be re-executed on the robot with high fidelity since the gripper's mechanical and sensing characteristics are unchanged.
            \end{itemize}
    \end{itemize}
    
    The entire design is 3D printable, offering a low-cost solution that can also be fabricated using metallic parts for enhanced reliability. We have open-sourced this design for the research community's use, with an iOS app available for free download at \url{https://apps.apple.com/cn/app/magiclaw/id6661033548}, along with accompanying documentation.

    \paragraph{Parallel Four-Bar as the Driving Mechanism} 
    Although there is no universally ``better'' choice of design for the drive mechanism, the parallel four-bar design offers several key advantages that may be suitable for this dual-purpose application: 
    \begin{itemize}
        \item \textbf{Field Use at a Low Cost}: Since its mechanism is driven, no rails are needed, which is great for small robots or end-effectors with tight mass budgets. The use of rolling joints means that it's less sensitive to dust, chips, or sprays than sliding guides, which is also 3D-printing-friendly.
        \item \textbf{A Big Stroke in a Thin Package}: The mechanism itself deploys and folds during usage, making it a more compact solution for field use while being capable of dealing with large-width objects during manipulation, covering most object sizes in daily life or even industrial scenarios.
        \item \textbf{Tunable Force Curve \& Compliance}: The jaw speed and force changes with the angular input, meaning that the motion is an arc (the fingers' orientation stays parallel but their paths aren't perfectly straight), which provides a high mechanical advantage near full closure for strong holding with a small motor.
    \end{itemize}

    \paragraph{Backdrivable Actuation for Tunable Interaction} 
    Each finger is driven by a small motor with an encoder for accurate position feedback. The gearing ratio is tuned to deliver sufficient gripping force for everyday objects while preserving \emph{back-drivability}, allowing the system to sense external contact forces and accommodate unmodeled variations. This design choice is especially helpful when switching between human-held demonstrations (where the user demands a responsive device) and robotic operation (where compliance prevents accidental damage to objects or the environment).

    \paragraph{Adaptive Fingertips with Omni-directional Perception}
    At the distal end of each four-bar linkage lies a \emph{Soft Polyhedral Network} (SPN) \cite{liu2024proprioceptive}, seen in Fig.~\ref{fig:MagiClaw}C. These fingertips have a flexible lattice pattern (e.g., a TPU-based 3D-printed mesh) that can \emph{conform in omni-directions} around diverse object profiles. Unlike silicone or purely elastic membranes, this lattice architecture provides both structural integrity and localized deformation nodes, thereby enhancing grip stability on irregular or deformable items.

    Inside each fingertip, we embed a miniature camera (e.g., a wide-angle micro camera) that observes the lattice from within. As external forces act on the SPN, the lattice geometry distorts. By tracking the shifting pattern of these lattice elements in real-time, a lightweight neural network infers \emph{6D force/torque} at the fingertip. Compared to conventional force sensors:
    \begin{itemize}
        \item \textbf{Low Cost}: The hardware cost is dominated by commodity micro cameras rather than specialized force-torque transducers.
        \item \textbf{High Spatial Resolution}: Deformation is recorded across the entire fingertip, reflecting \emph{where} and \emph{how} contact forces are applied.
        \item \textbf{Minimal Additional Bulk}: The sensing mechanism is entirely contained within the existing soft structure, maintaining a low overall profile.
    \end{itemize}

    \begin{figure}[t]
            \centering
            \includegraphics[width=1\linewidth]{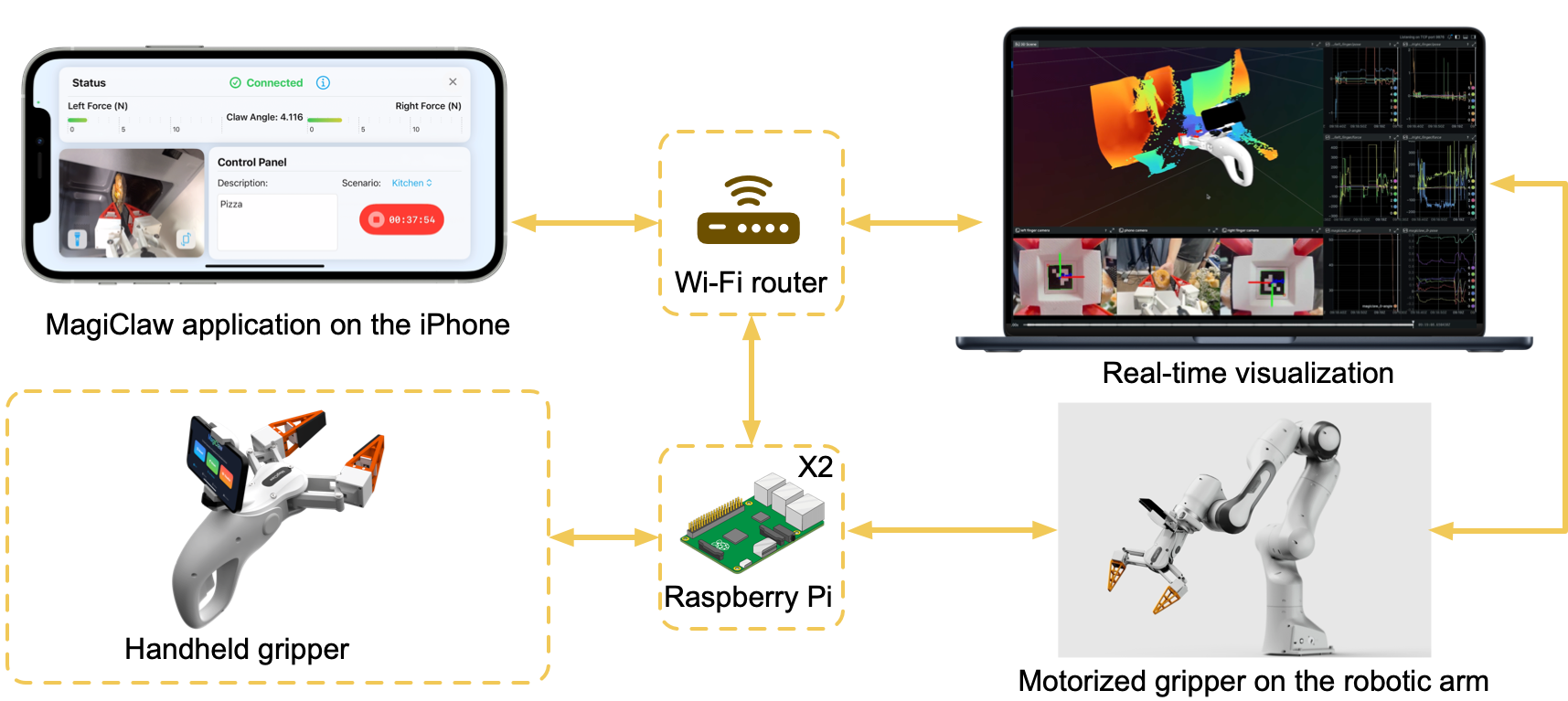}
            \caption{\textbf{Data visualization interface (via Rerun) and communication topology.} Multi-modal data streams (pose, depth, fingertip images, etc.) are synchronized across devices for real-time or offline analysis.}
            \label{fig:Communication}
    \end{figure}

    \paragraph{Smartphone Sensing for Robotics}
    A hallmark of MagiClaw is the integration of a consumer smartphone, specifically an iPhone Pro (although other smartphone brands or series may offer similar functionalities, depending on their hardware capabilities), into the gripper assembly (Figs.~\ref{fig:MagiClaw}A\&B). Currently, this smartphone offers the following capabilities. By leveraging off-the-shelf smartphone hardware, MagiClaw benefits from on-device processing capabilities, integrated communication (Wi-Fi, Bluetooth, cellular), and a user-friendly interface for real-time monitoring.
    \begin{itemize}
        \item \textbf{LiDAR Depth Sensing}: High-frame-rate 3D mapping of the environment, enabling real-time object detection, scene reconstruction, or augmented-reality overlays.
        \item \textbf{RGB Video}: High-resolution images or videos from the rear camera for visual context, teleoperation views, or training data.
        \item \textbf{Gripper Pose Tracking}: Using Apple's ARKit framework and provided APIs to track the gripper's orientation and movement relative to the global frame \cite{apple2025arkit}.
        \item \textbf{Audio Capture}: Potentially useful for event detection (e.g., collision sounds or object rattles).
    \end{itemize}

    \paragraph{Motor Drivers and Microcontroller}
    Beyond the smartphone, a microcontroller (such as the Raspberry Pi 5) handles low-level tasks. This architecture isolates time-critical control from higher-level processes on the smartphone, maintaining robust performance despite the smartphone's variable compute load. It 1) receives setpoints (e.g., desired grip width) and commands the servo or DC motor drivers accordingly, 2) reads encoder values to relay finger positions and detect contact or stalling conditions, 3) measures interaction forces and transmits them to the handheld gripper, enabling haptic feedback for the user, and 4) coordinates timestamping of finger data with the smartphone's sensor streams.

    \paragraph{System Communication Architecture}
    The communication architecture of the MagiClaw system is illustrated in Fig.~\ref{fig:Communication}. An iPhone, mounted on the hand-held or motorized gripper, connects to a Wi-Fi router over a wireless network. Both the handheld and motorized grippers contain a motor, each wired to a Raspberry Pi. These two Raspberry Pi boards communicate with their respective motors via the CAN protocol.
    
    Each gripper also integrates two SPNs, each equipped with an internal camera. The cameras connect to the Raspberry Pi using Wi-Fi hotspots that they create themselves. The Raspberry Pis, in turn, communicate wirelessly with the central Wi-Fi router.
    
    A computer running Rerun for real-time data visualization is also connected to the same router. It receives data streams from all devices in the local network. The MagiClaw app, running on the iPhone, serves a dual purpose. First, it displays 6D force/torque data from the SPN. Second, it broadcasts its own 6D pose, RGB images, and depth images, which are accessible to all devices within the local network. Additionally, the app offers direct control of the gripper motors, allowing for start and stop commands.
	
\section{Experimental Validation: The Imitation Game}
\label{sec:result}
    
    We validate the MagiClaw system through a series of use cases framed as \textit{The Imitation Game}—a spectrum of tasks that demonstrate the system's capacity to capture holistic human actions and transfer them to a robot. These experiments serve to confirm the utility of our unified hardware and multi-modal sensing approach.

\subsection{High-Fidelity Teleoperation and Immersive Demonstration}

\begin{figure}[b]
        \centering
        \includegraphics[width=\linewidth]{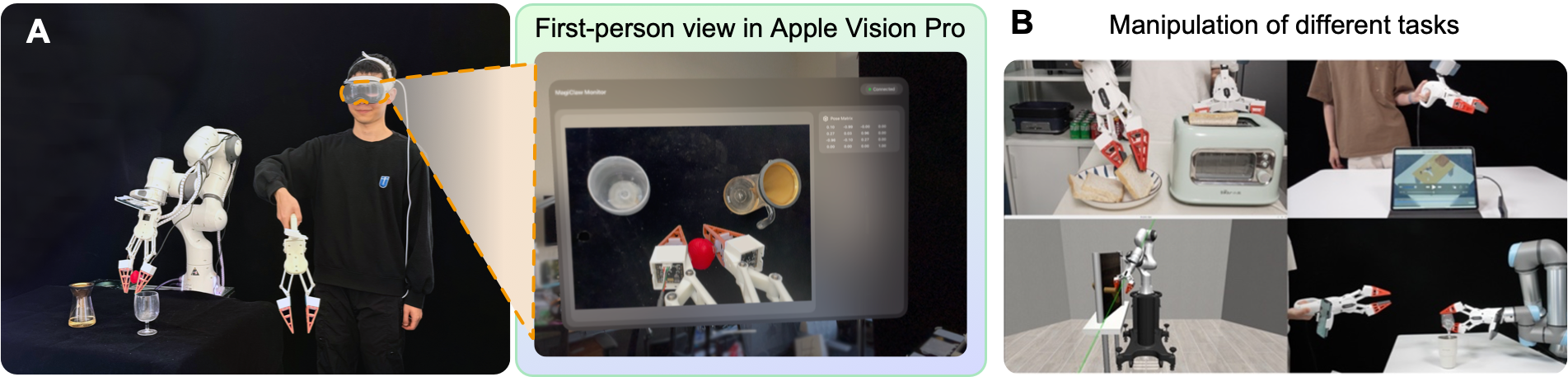}
        \caption{\textbf{MagiClaw in Action.} (A) An operator performs teleoperation with an immersive, first-person view provided by an Apple Vision Pro headset, which overlays real-time sensor data. (B) The system's versatility is demonstrated across various manipulation tasks, both in the real world and in simulation, showcasing its adaptability.}
        \label{fig:Tasks}
    \end{figure}

    A primary use case for MagiClaw is real-time teleoperation, where a human operator's actions are mirrored by a robot-mounted gripper (Fig.~\ref{fig:Tasks}). 
    The operator uses a hand-held MagiClaw, and its state (6D pose and grip width) is streamed to the robot. This setup validates several key system capabilities:
    \begin{itemize}
        \item \textbf{Intuitive Control and Data Capture:} The direct physical interface allows for natural and dexterous manipulation. Simultaneously, the system logs a complete, synchronized dataset of the operator's actions and the resulting environmental interactions.
        \item \textbf{Closed-Loop Force Feedback:} The vision-based force estimation from the robot's SPN fingertips is streamed back to the operator's hand-held device, providing haptic feedback. This allows the operator to "feel" the interaction forces, enabling delicate tasks that would be impossible with visual feedback alone.
    \end{itemize}
    
    To further enhance the operator's situational awareness, we integrate this system with an \textbf{Apple Vision Pro} mixed-reality headset (Fig.~\ref{fig:Tasks}A). The headset provides a first-person view from the robot's perspective, overlaying real-time sensor data (e.g., force vectors, depth maps) to enhance situational awareness. This immersive interface significantly reduces the cognitive load on the operator, enabling the demonstration of highly precise and complex maneuvers. This capability validates our claim of creating a user-friendly and powerful interface for demonstration.

\subsection{Learning from Multi-Modal Replays}

\begin{figure}[h]
        \centering
        \includegraphics[width=\linewidth]{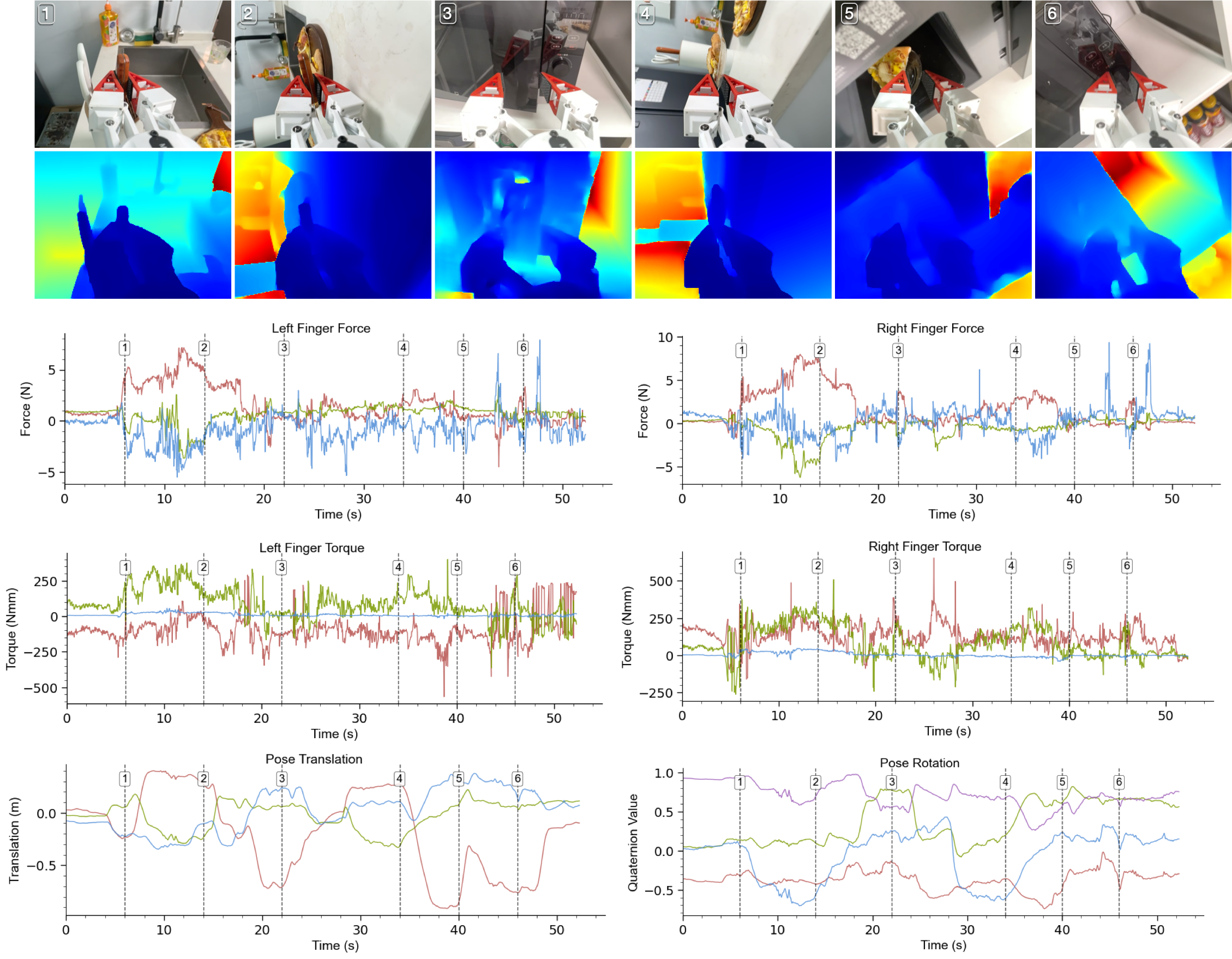}
        \caption{\textbf{Collected Multi-Modal Data.} Task of transferring a slice of pizza to a plate and heating it in the microwave, showing RGB and depth images alongside temporal variations of left and right fingertip 6D force/torque and MagiClaw gripper's 6D pose.}
        \label{fig:Data}
    \end{figure}

    The rich datasets collected during teleoperation or offline hand-held demonstrations (Fig.~\ref{fig:Data}) form the foundation for policy learning. This workflow validates the core hypothesis that unified hardware reduces the domain gap. Please refer to the Supplementary Video for further demonstration.
    \begin{itemize}
        \item \textbf{Direct Policy Transfer:} Because the demonstration and deployment hardware are identical, simple behavioral cloning policies can be trained on the collected data and directly deployed on the robot with minimal performance degradation from sensor or kinematic mismatch.
        \item \textbf{Seeding for Advanced Learning Algorithms:} The multi-modal data is ideally suited for training more sophisticated models. For example, synchronized force and visual data can be utilized in offline reinforcement learning to learn reward functions that encourage gentle contact, or to train predictive models that anticipate contact events based on visual input.
    \end{itemize}

    The ability to replay demonstrations on the physical robot allows for iterative debugging and policy refinement. Discrepancies between the original demonstration and the robotic execution can be logged and used to further improve the learned model. 

\subsection{Validation in Contact-Rich Scenarios}

    We further validate MagiClaw's utility in advanced tasks where force and tactile feedback are critical. Please refer to the Supplementary Video for further demonstration.
    
    These scenarios confirm that the high-fidelity contact data captured by MagiClaw enables the learning of skills that are beyond the reach of systems relying solely on kinematic and visual information. 

\section{Conclusion}
\label{sec:conclusion}

    This paper introduced \textit{MagiClaw}, a multi-modal gripper system designed to accelerate research in robot learning by directly addressing the demonstration-to-deployment gap. Its novel dual-purpose design, which unifies the hardware for data collection and policy execution, fundamentally minimizes domain shift. By integrating vision-based proprioceptive force sensing in soft fingertips with comprehensive exteroceptive sensing capabilities from a commodity smartphone, MagiClaw provides a low-cost yet powerful turnkey solution for generating rich, contact-centric datasets.
    
    We have demonstrated how this integrated system enables high-fidelity teleoperation with immersive feedback, streamlines data collection for policy learning, and proves effective in challenging, contact-rich tasks. By simplifying and democratizing access to holistic action data, we believe MagiClaw will catalyze progress in data-driven robotics, paving the way for more dexterous, adaptable, and human-like manipulation.
    
    \textbf{Limitations and Future Work.} The current system depends on low-latency wireless communication, which may bottleneck in congested networks. Vision-based force estimation is cost-effective but requires fingertip-specific calibration and training, suggesting room for streamlining. Reflective surfaces challenge LiDAR depth accuracy. As the iPhone is not a hard real-time system, iOS scheduling and thermal throttling limit the safety-critical control, making the integration of off-the-shelf cooling solutions a preferable option.
    
    Our future work will focus on improving the system's robustness and expanding its capabilities. We plan to explore onboard policy learning directly on the integrated smartphone, investigate more sample-efficient calibration methods for the SPN fingertips, and develop a library of pre-trained models for common manipulation tasks. Crucially, we intend to open-source the hardware designs and core software modules to foster collaboration and empower the wider robotics research community to build upon our work.

\section*{Supplementary Materials}
\label{sec:SupMat}

    Please refer to this link: \url{https://bionicdl.ancorasir.com/?p=2162} for a supplementary video of this paper, the iOS app, and the full documentation.

\bibliographystyle{unsrt}
\bibliography{References}  
\end{document}